\documentclass[10pt,twocolumn,letterpaper]{article}

\usepackage{iccv}
\usepackage{times}
\usepackage{epsfig}
\usepackage{graphicx}
\usepackage{amsmath}
\usepackage{amssymb}
\usepackage{color}
\usepackage{tabularx}
\usepackage{multirow}
\usepackage{hhline}
\usepackage{subfig}
\usepackage{booktabs}
\usepackage{epstopdf}
\usepackage{amsmath}
\usepackage{url}
\usepackage{epstopdf}

\newcommand{\ignore}[1]{}

\def\etal{\emph{et al}\onedot}

\DeclareMathOperator*{\argmax}{arg\,max}

\iccvfinalcopy 

\def\iccvPaperID{225} 

\ificcvfinal\fi
\begin{document}

\title{Guiding Long-Short Term Memory for Image Caption Generation}

\author{Xu Jia \\
KU Leuven ESAT-PSI, iMinds \\
{\tt\small Xu.Jia@esat.kuleuven.be}
\and
Efstratios Gavves\thanks{This work was carried out while he was in KU Leuven ESAT-PSI.}\\
QUVA, University of Amsterdam\\
{\tt\small E.Gavves@uva.nl}
\and
Basura Fernando \\
ACRV, The Australian National University \\
{\tt\small basura.fernando@anu.edu.au}
\and
Tinne Tuytelaars \\
KU Leuven ESAT-PSI, iMinds \\
{\tt\small Tinne.Tuytelaars@esat.kuleuven.be}
}

\maketitle

\begin{abstract}
In this work we focus on the problem of image caption generation.
We propose an extension of the long short term memory (LSTM)  model,
which we coin gLSTM for short.
In particular, we add semantic information extracted from the image
as extra input to each unit of the LSTM block,
with the aim of guiding the model towards solutions 
that are more tightly coupled to the image content.
Additionally, we explore different length normalization strategies for beam search
in order to prevent from favoring short sentences.
On various benchmark datasets such as Flickr8K, Flickr30K and MS COCO, we obtain results that are on par with
or even outperform the current state-of-the-art.
\end{abstract}

\section{Introduction}
Recent successes in visual classification have shifted the interest of the community
towards higher-level, more complicated tasks,
such as image caption generation~\cite{FarhadiECCV10, YangEMNLP11, KulkarniPAMI13, MitchellEACL12,
KuznetsovaACL12, KuznetsovaACL13, KuznetsovaTACL14, MasonACL14,
KirosICML14,Mao14,Karpathy14,Vinyals14,Donahue14,Xu15}.
Although for a human describing a picture is natural,
it is quite difficult for a computer to imitate this task.
It requires the computer to have some level of semantic understanding of the content of an image,
including which kinds of objects there are, how they look like, what they are doing, and so on.
Last but not least, this semantic understanding has to be structured into a human-like sentence.

Inspired by recent advances in machine translation~\cite{ChoEMNLP14, Bahdanau14, SutskeverNIPS14},
neural machine translation models have lately been applied to the image
caption generation task~\cite{Mao14,Karpathy14,Vinyals14,Donahue14,Xu15}, with remarkable success.
In particular, compared to template-based methods, that use a rigid
sentence structure, and transfer-based methods, that re-use
descriptions available in the training data, methods based on
neural machine translation models stand out thanks to their
capability to generate new sentences.
They manage to effectively generalize beyond the sentences seen at training time,
which is possible thanks to the language model learnt.
\begin{figure}
 \centering
\includegraphics[width=1.0\linewidth]{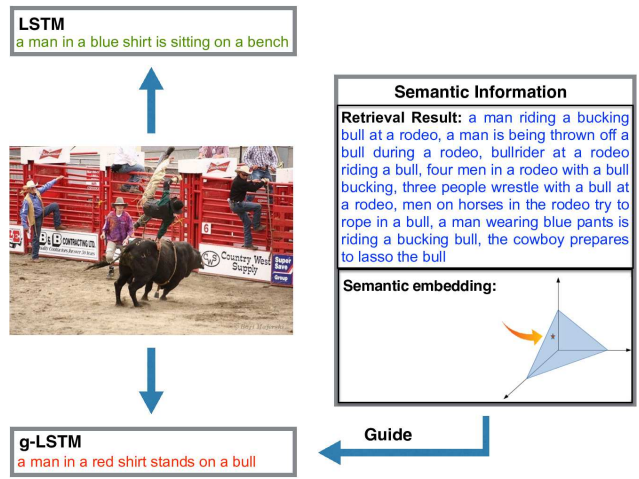}
\caption{Image caption generation using LSTM and the proposed gLSTM.
The generation by LSTM and gLSTM and the cross-modal result that is used as guidance,
are marked respectively in green, red and blue.} 
\label{fig:example_page_1}
\vspace{-5mm}
\end{figure}
Most neural machine translation models follow an encoder-decoder pipeline~\cite{ChoEMNLP14, Bahdanau14, SutskeverNIPS14},
where the sentence in the source language is first encoded into a fixed-length embedding vector,
which is then decoded to generate a new sentence in the target language.
For machine translation, parallel corpora are 
typically used for learning and evaluating the model~\cite{ChoEMNLP14, Bahdanau14, SutskeverNIPS14}.
The pairs of sentences in the source and target languages usually share similar sentence structures
(often including regular phrases and the same order of words). This structural information
is encoded in the fixed-length embedding vector and is helpful to the translation.

Applied to caption generation, the aim is to ``translate'' an image into a sentence describing it.
However, it is questionable whether these models can cope with the large differences between the two modalities.
The structure of the visual information is very different from the structure of the description to be generated.
During the encoding phase, the algorithm compresses all visual information into an embedding vector.
Yet this vector is unlikely to capture the same level of structural information needed for
correctly generating the textual description in the subsequent decoding phase.

One of the latest state-of-the-art methods~\cite{Vinyals14}
uses a convolutional neural network (CNN) for the encoding step and
long-short term memory (LSTM) network for the decoding step.
We notice that sometimes the generated sentence seems to ``drift away''
or ``lose track'' of the original image content, 
generating a description that is common in the dataset, 
yet only weakly coupled to the input image. 
We hypothesize this is because the decoding step needs to find a balance between two, sometimes contradicting,
forces: on the one hand, the sentence to be generated needs to describe the image content;
on the other hand, the generated sentence needs to fit the language model, with more likely word combinations to be preferred.
The system then may ``lose track'' of the original image content if the latter force starts to dominate.
From an image caption generation point of view, however, staying close to the image content may
be considered the most important of the two.

To overcome the limitation of the basic encoding-decoding pipeline,
extended pipelines have been proposed in the context of 
both machine translation~\cite{Bahdanau14} and image caption generation~\cite{Xu15}. 
They introduce an attention mechanism to align the information in both the source and target domains,
so that the model is able to attend to the most relevant part 
in the sentence from the source language or image.

Here, we propose an alternative extension of the LSTM model, that works at a more global scale.
We start by extracting semantic information from the image and then use it to ``guide'' the decoder,
keeping it ``on track'' by adding a positive bias to words that are semantically linked to the image content.
More specifically, we add semantic information as an extra input to the gate of each LSTM memory cell.
This extra input can take many different forms 
as long as they build a semantic connection between the image and its description.
\eg a semantic embedding, a classification or retrieval result.
As an illustration we experiment with features either obtained
from a multimodal semantic embedding using CCA or, 
the cross-modal retrieval results based on that semantic embedding.

Our contributions are two-folded.
As our main contribution, we present an extension of LSTM which is guided by semantic information of image.
We coin a term gLSTM for the proposed method.
We show experimentally on multiple datasets that such guiding is beneficial for learning to generate image captions.
As an additional contribution, we make the observation that current inference methodologies for caption generation
are heavily biased towards short sentences.
We show experimentally that this hurts the quality of the generated sentences and
therefore propose sentence normalization, which further improves the results.
In the experiments, we show that the proposed method is
on par or even outperforms the latest published and unpublished state-of-the-art
on the popular datasets.

\vspace{-1mm}
\section{Related Work}


\noindent\textbf{Caption generation.} 
The literature on caption generation can be divided into three families.
In the first family we have template-based methods~\cite{FarhadiECCV10, YangEMNLP11, KulkarniPAMI13, MitchellEACL12}. 
These approaches first detect objects, actions, scenes and attributes,
then fill them in a fixed sentence template, \eg using a subject-verb-object template.
These methods are intuitive and can work with out-of-the-box visual classification components. 
However, they require explicit annotations for each class.
Given the typically small number of categories available, 
these methods do not generate rich enough captions. 
Moreover, as they use rigid templates the generated sentence is less natural.

The second family follows a transfer-based caption generation strategy
\cite{KuznetsovaACL12, KuznetsovaACL13, KuznetsovaTACL14, MasonACL14}.
They are related to the image retrieval. 
These methods first retrieve visually similar images, 
then transfer captions of those images to the query image. 
The advantage of these methods is that the generated captions 
are more human-like than the generations by template-based methods.
However, because they directly rely on retrieval results among training data, 
there is little flexibility for them to add or remove words based on the content of an image.

Inspired by the success of neural networks in machine translation ~\cite{ChoEMNLP14, Bahdanau14, SutskeverNIPS14}, 
recently people have proposed to use neural language models for caption generation. 
Instead of translating a sentence from a source language into a target one, 
the goal is to translate an image into a sentence that describes it. 
In~\cite{KirosICML14} a multimodal log-bilinear neural language model is 
proposed to model the probability distribution of a word conditioned on an image and previous words. 
Similarly, Mao \etal~\cite{Mao14} and Karpathy \etal~\cite{Karpathy14} have proposed to 
use a multimodal recurrent neural network~\cite{RNN} model for caption generation. 
Vinyals \etal\cite{Vinyals14} and Donahue \etal\cite{Donahue14} 
have proposed to use LSTM~\cite{LSTM}, an advanced Recurrent Neural Network for the same task.
Very recently, Xu \etal~\cite{Xu15} have proposed to integrate visual attention into the LSTM model 
in order to fix its gaze on different objects during the generation of corresponding words.
Neural language models have shown great prospects in generating human-like image captions.
Most of these methods follow a similar encoding-decoding framework, 
except for the very recent method~\cite{Xu15} 
which jointly learns visual attention and caption generation.
However, ~\cite{Xu15} requires location sampling both during training and testing, 
making the method more complicated.
While they focus more on local information, 
our method rather exploits global cues.

\noindent\textbf{Overview.} 
Our work belongs to the third family of caption generation methods 
which uses a neural language model to generate captions.
Different from the above methods, however, 
we propose to make use of the semantic information to guide the generation and 
propose an extension of LSTM model, coined gLSTM
for the use of semantic information.
The semantic information here denotes the correlation between image and its description,
which is obtained in a similar manner as in transfer-based methods.
Experiments illustrate that semantic information brings significant improvement in the performance 
and our model outperforms recently proposed state-of-the-art methods~\cite{Karpathy14, Vinyals14}. 
Interestingly, the proposed model is able to perform on par with the latest and unpublished state-of-the-art~\cite{Xu15}, 
despite their use of more complicated models that require location sampling during training and test stage.

\vspace{-1mm}
\section{Background}
\subsection{The LSTM Model}

A Recurrent Neural Network (RNN) is a good choice to model temporal dynamics in sequences.
However, it is difficult for traditional RNN to learn long-term dynamics 
because of the issue of vanishing and exploding gradients~\cite{LSTM}.
The Long Short-Term Memory (LSTM) network is proposed in~\cite{LSTM} to address these issues.
The core of the LSTM architecture is the memory cell, which stores the state over time, and
the gates, which control when and how to update the cell's state.
There are many variants with different connections between the memory cell and the gates.

\begin{figure}[t!]
\centering
\includegraphics[width=\linewidth]{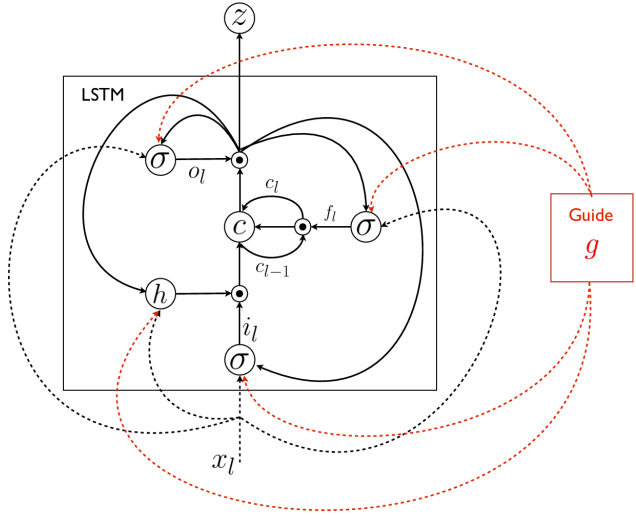}
\caption{The LSTM block in black, the proposed gLSTM network in black and red. 
Striped lines stand for external connections. 
By considering semantic information as an extra input,  
we encourage the network to refresh its memory following a global guide.}
\label{fig:glstm}
\vspace{-5mm}
\end{figure}

The LSTM block that our model is built on follows the \textit{LSTM with No Peepholes} architecture~\cite{Greff15},
which is illustrated in Figure~\ref{fig:glstm} in black. 
The memory cell and gates in an LSTM block are defined as follows:
\begin{eqnarray}
\vspace{-5mm}
i_l & = & \sigma (W_{ix} x_l + W_{im} m_{l-1}) \\
f_l & = & \sigma (W_{fx} x_l + W_{fm} m_{l-1}) \\
o_l & = & \sigma (W_{ox} x_l + W_{om} m_{l-1}) \\
c_l & = & f_l \odot c_{l-1} + i_l \odot h(W_{cx} x_l + W_{cm} m_{l-1}) \\
m_l & = & o_l \odot c_l \label{lstm-memory} 
\vspace{-5mm}
\end{eqnarray}
where $\odot$ represents the element-wise multiplication, $\sigma(\cdot)$ represents the sigmoid function and 
$h(\cdot)$ represents the hyperbolic tangent function.
The variable $i_l$ stands for the input gate , $f_l$ for the forget gate, $o_l$ for the output gate of the LSTM cell, 
$c_l$ is the state of the memory cell unit and 
$m_l$ is the hidden state, that is the output of the block generated by the cell.
The variable $x_l$ is the element of the sequence at timestep $l$ and
$W_{[\cdot][\cdot]}$ denote the parameters of the model.
%
%

\subsection{Caption Generation with LSTM}
\label{sec:lstm_generation}
The pipeline for caption generation with the RNN model~\cite{Mao14, Karpathy14, Vinyals14, Donahue14, Xu15} is 
inspired by the encoder-decoder principle in Neural Machine Translation~\cite{ChoEMNLP14, SutskeverNIPS14, Bahdanau14}.
An encoder is used to map a variable length sequence in the source language into a distributed vector
and a decoder is used to generate a new sequence in the target language conditioned on this vector.
During training, the goal is to maximize the log-likelihood of correct translation 
given the sentence in the source language.
When applying this principle to caption generation, 
the goal becomes to maximize the log-likelihood of the image caption given an image, namely
\begin{equation}
\vspace{-2mm}
\argmax_{\theta} \sum_i \log p(s_{1:L^i}^i | x^i, \theta),
\label{loglikelihood}
\vspace{-1mm}
\end{equation}
where $x^i$ denotes an image, $s_{1:L^i}^i$ denotes a sequence of words in a sentence of length ${L^i}$ and 
$\theta$ denotes the model parameters.
For simplicity, in the following part we drop the superscript $i$ whenever it is clear from the context. 
Since each sentence is composed of a sequence of words, 
it is natural to use Bayes chain rule to decompose the likelihood of a sentence,
\begin{equation}
\vspace{-2mm}
\log p(s_{1:L}| x, \theta) = \log p(s_1 | x, \theta) + \sum_{l=2}^L \log p(s_l | x, s_{1:l-1}, \theta),
\label{log-sentence}
\vspace{-1mm}
\end{equation}
where $s_{1:l}$ stands for the part of the sentence up to the $l$-th word. 
To maximize the objective in eq.~\eqref{loglikelihood} over the whole training corpus, 
we need to define the log-likelihood $\log p(s_l | x, s_{1:l-1}, \theta)$,
which can be modeled with the hidden state of a timestep in RNN.
The probability distribution of the word at timestep $l+1$ over the whole vocabulary 
is computed using the softmax function $z(\cdot)$ based only on the output $m_l$ of the memory cell,
$p_{l+1} = z(m_l)$ similar to~\cite{Vinyals14}.

To feed images and sentences to LSTM,
they need to be encoded as fixed-length vectors.
For the image, CNN features are first computed 
and then mapped to an embedding space via a linear transformation.
For the sentence, each word is first represented as a one hot vector and 
then mapped to the same embedding space via a word embedding matrix.
Finally, an image and sequence of words in a sentence are concatenated to form a new sequence,
that is, the image is treated as the beginning symbol of the sequence and 
the sequence of words forms the remaining part of the new sequence.
This sequence is fed to the LSTM network for training by iterating the recurrence connection for $l$ from $1$ to $L^i$.
The parameters of the model include the linear transformation matrix for image features, 
the word embedding matrix and the parameters of LSTM.

\subsection{Normalized Canonical Correlation Analysis}
\label{sec:normalized_cca}
To build our semantic representation, we rely on normalized Canonical Correlation Analysis (CCA), 
proposed in~\cite{GongKIL14} to address the cross-modal retrieval problem.
Canonical correlation analysis (CCA)~\cite{HotellingCCA} is a popular method 
used to map visual and textual features into a common semantic space.
CCA aims at learning projection matrices $U_1$ and $U_2$ for two views $X_1$ and $X_2$ 
such that their projections are maximally correlated, namely,
\begin{equation}
\vspace{-2mm}
 \argmax_{U_1,U_2} \frac{{U_1} {\Sigma}_{X_1 X_2} {U_2}}{\sqrt{{U_1} {\Sigma}_{X_1 X_1} {U_1}} \sqrt{{U_2} {\Sigma}_{X_2 X_2} {U_2}}},
\end{equation}
where ${\Sigma}_{X_1 X_2}$, ${\Sigma}_{X_1 X_1}$ and ${\Sigma}_{X_2 X_2}$ are the covariance matrices.
The CCA objective function can be solved via generalized eigenvalue decomposition.
The normalized CCA is computed by using a power of the eigenvalues 
to weight the corresponding columns of the CCA projection matrices,
and followed by L2 normalization, that is,
\begin{equation}
\vspace{-2mm}
 g_1 = \frac{X_1 U_1 D^p}{\lVert {X_1 U_1 D^p} \rVert}, \quad
 g_2 = \frac{X_2 U_2 D^p}{\lVert {X_2 U_2 D^p} \rVert}
 \label{cca}
\end{equation}
where $D$ is a diagonal matrix whose elements are set to the eigenvalues of corresponding dimensions,
while $g_1$ and $g_2$ denote the semantic representation of the two views.
Cosine similarity is used to find the nearest neighbor in the learned common semantic space~\cite{GongKIL14}.

\vspace{-1mm}
\section{The Proposed Methods}

In this section, we describe the proposed extension of the LSTM model for the caption generation task.
In the new architecture, we add semantic information to the computation of the gates and cell state.
The semantic information here is extracted from images and their descriptions,
serving as a guide in the process of word sequence generation.

\subsection{gLSTM}
The generation of a word in the LSTM model mainly depends on the word embedding at the current timstep and the previous hidden state 
(which includes image information at the beginning).
This process goes step by step until it encounters the end token of a sentence.
However, as this process continues, the role of the image information, which is only fed at the beginning, 
becomes weaker and weaker.
Words generated at the beginning of a sequence also suffer from the same problem.
Therefore, for a long sentence, it may carry out the generation almost ``blindly'' towards the end of the sentence.
Though LSTM is able to keep long-term memory to some extent, 
still it poses a challenge for sentence generation~\cite{Cho14, Bahdanau14}.
In the proposed model, the generation of words is carried out 
under the guidance of global semantic information.
Our extension of LSTM model is named gLSTM.
The memory cell and gates in a gLSTM block are defined as follows:
\begin{eqnarray}
\vspace{-3mm}
i_l' & = & \sigma (W_{ix} x_l + W_{im} m_{l-1} + W_{iq} g) \label{glstm-input} \\
f_l' & = & \sigma (W_{fx} x_l + W_{fm} m_{l-1} + W_{fq} g) \\
o_l' & = & \sigma (W_{ox} x_l + W_{om} m_{l-1} + W_{oq} g) \\
c_l' & = & f_l' \odot c_{l-1}' + i_l' \odot h(W_{cx} x_l + \nonumber \\
    &   & + W_{cm} m_{l-1} + W_{cq} g) \\
m_l & = & o_l' \odot c_l' 
\label{glstm-memory}
\vspace{-3mm}
\end{eqnarray}
where $g$ denotes the vector representation of semantic information.
Compared to the standard LSTM architecture, in gLSTM
we add a new term to the computation of each gate and cell state.
This new term represents the semantic information 
which works as a bridge between visual and textual domains.
The semantic information $g$ does not depend on the timestep $l$, 
hence working as a global guide during the caption generation.
The guidance term can also be timestep dependent in expense of higher complexity models.
We summarize with red the gLSTM network architecture additions in Figure~\ref{fig:glstm}.

\subsection{Semantic Information.}
\label{sec:contextual_info}
In this section, we will detail several kinds of semantic information 
that can be used as guidance in our model.
Intuitively, there are three ways to extract the semantic information.
First, we treat it as a cross-modal retrieval task and simply use the retrieved sentences as semantic information.
Alternatively, semantic information can also be represented as the embedding in a semantic space 
where visual and textual representations are equivalent.
The last one is to use the image itself as guidance.

\noindent\textbf{Retrieval-based guidance (ret-gLSTM).} 
The retrieval-based guidance is inspired by the transfer-based caption generation methods.
Though the generated sentence given by transfer-based methods may not be totally correct,
they do have something in common with the true captions annotated by humans.
Given an image, we first do the cross-modal retrieval whose aim is to find relevant texts to the query image.
Then we collect descriptions with top rankings.
Instead of generating a sentence by making direct modification on these sentences,
we treat these captions as auxiliary information and 
feed them to the neural language model we propose in the previous section.
These sentences may not match perfectly to the image. 
However, they provide rich semantic information for the image.
Since these sentences are annotated by humans, 
the words in these sentences are very natural and 
have a high probability to appear in the reference captions.

The cross-modal retrieval method used here is based on the normalized CCA mentioned in Section~\ref{sec:normalized_cca}.
In this paper, image and text features correspond to the two views for CCA.
CNN features are computed for the images and TF-IDF weighted BoW features are computed for the sentences.
We project both images and sentences from their own domain to the common semantic space.
Given an image query, the closest sentences are then retrieved based on cosine similarity.
We select the top $T$ retrieved sentences from the training set ($T=15$ in this paper).
These sentences are represented by a bag-of-words (BoW) vector 
which is fed as extra input, i.e. the guide to the gLSTM model.

\noindent\textbf{Semantic embedding guidance (emb-gLSTM).}  
We can explicitly use the result of cross-modal retrieval as guidance as mentioned above.
We can also implicitly use the intermediate result of cross-modal retrieval,
that is the semantic representation computed using normalized CCA as the extra input.
An image is mapped into the common semantic space by the learned projection matrix
and the computed semantic embedding is fed to gLSTM model as the guide.
It is assumed that in the common semantic space of CCA both views share equivalent embedding representation.
Therefore, we can treat the projected representation from image domain as equal to the one projected from text domain.
Compared to ret-gLSTM model, 
the semantic representation has much lower dimensionality than the BoW representation 
and saves the computation of finding nearest neighbors.
In addition, we also find it even performs better than the previous method.


\noindent\textbf{Image as guidance (img-gLSTM).} 
Finally, we experiment with the image itself as the extra input.
This is motivated by the fact that CCA is a linear transformation. 
A natural question then is 
whether we can learn this projection matrix directly during the training of the gLSTM model.
Therefore, we add the image itself as a third kind of extra input.
We experimentally verify this by simply feeding the image feature itself to the gLSTM model, namely $g=x$, and 
let the network learn the semantic information from scratch. 

\subsection{Beam Search with Length Normalization}
In the generation stage, with a vocabulary of size $K$, 
there are $K^l$ sentences of length $l$ as potential candidates for an image caption, where $l$ is unknown. 
Ideally, we want to find the sentence, which maximizes the log-likelihood of eq.~\eqref{log-sentence}. 
Considering the exponential search space, however, exhaustive search is intractable. 
Therefore, a heuristic search strategy is employed instead.

Here we use \emph{beam search}, which is a fast and effective decoding method 
for RNN-based models~\cite{Graves12, SutskeverNIPS14}.
At each iteration only the $T$ hypotheses generations with the highest log-likelihood are kept in the beam pool.
The search along one beam stops 
once it encounters an end-of-sequence token which is generated given previous words along the beam.
The searching process continues until the searching along all beams in the pool stops.

It is problematic to directly use the log-likelihood of words as the criterion to select a generation. 
Since the log-likelihood of each single word is negative (because the probability is smaller than 1),
summation over the log-likelihood of more words lead to a smaller value.
Therefore, when the beam width is larger than 1, 
the beam search stops early is more likely to be selected as the final caption,
regardless of the quality of each generated word in the beam.
That means this kind of beam search favors shorter sentence, 
which is also observed in ~\cite{Graves13, Cho14}

Interestingly, the bias towards short sentences tends to favor the low order of BLEU scores (BLEU@1,2), 
commonly used to evaluate machine translation algorithms. 
Hence, short sentences not only tend to dominate the inference, 
but also obscure the evaluations and methodology comparisons. 
To remedy the bias towards short sentences during inference, 
we propose to normalize the log-likelihood of words by length, namely
\begin{equation}
p = \frac{1}{\Omega(\ell)}\sum_{l=1}^{\ell} \log p(s_l | x, s_{1:l}, \theta)
\label{eq:log-sentence-norm}
\end{equation}
We investigate various forms for $\Omega$ to do the normalization.

{\flushleft{\textbf{Polynomial normalization.} }}
A first possibility is to set $\Omega(\ell)=|\ell|^m$.
Notice that when $m=1$, eq.~(\ref{eq:log-sentence-norm}) becomes the definition of the perplexity.
We use $m=1$ in our paper.
This kind of normalization punishes short sentences.

{\flushleft{\textbf{Min-hinge normalization.} }}
Intuitively we want to automatically generate a sentence whose length is close to the ground truth.
Since in the test stage we do not know the length in advance,
we use the average length of the sentences in the training data as a reference.
We define the min-hinge length function as $\Omega(\ell)=\min\{\ell, \mu\}$. 
This means a generated sentence is only punished 
when it is shorter than the average length $\mu$.
For sentences that are long enough,
we only pay attention to their log-likelihood.

{\flushleft{\textbf{Max-hinge normalization.} }}
Similarly, we define the max-hinge length function , $\Omega(\ell)=\max\{\ell, \mu\}$. 
Instead of penalizing short sentences, the max-hinge function favors long sentences.

{\flushleft{\textbf{Gaussian normalization.} }}
We can also employ a Gaussian function, $\Omega(\ell) \sim \mathcal{N}(\mu, \sigma)$ to normalize the loglikelihood, 
where the $\mu$ and $\sigma$ are the mean and the standard deviation of the sentence lengths in the training corpus. 
The Gaussian regularization encourages the inference to select sentences 
that have similar lengths as the sentences in the training set.

We experimentally verify the effectiveness of these strategies in Section~\ref{sec:exp_norm}.

\vspace{-1mm}
\section{Experiments}
\noindent\textbf{Datasets and experimental setup.} We perform experiments on the following datasets.

\noindent\textit{Flickr8k~\cite{flickr8k}, Flickr30k~\cite{flickr30k} and MS COCO~\cite{mscoco}}. 
The Flickr8k dataset is a popular dataset composed of $8,000$ images in total collected from Flickr, 
divided into a training, validation and test set of $6,000$, $1,000$ and $1,000$ images respectively. 
Each image in the dataset is accompanied with 5 reference captions annotated by humans. 
Similar to Flickr8k, the Flickr30k dataset contains $31,000$ images collected from Flickr, 
together with 5 reference sentences provided by human annotators. 
However, it does not provide a split setting file.
So we use the publicly available split setting used in~\cite{KarpathyNIPS14} and ~\cite{Karpathy14}, 
that is, $29,000$ images for training, $1,000$ for validation and $1,000$ for testing. 
The large scale dataset MSCOCO contains 82,783 images for training and 40504 for validation, 
with each image associated with 5 captions. 
Note that we donot evaluate it on the test set used for MS COCO Image Captioning challenge 
but use the publicly available splits used in previous work~\cite{Karpathy14}, 
that is, all 82,783 images from training set for training and 5,000 images from validation set for validation and testing.

\begin{figure*}[t!]
\vspace{-5mm}
\centering
\includegraphics[width=1.0\textwidth]{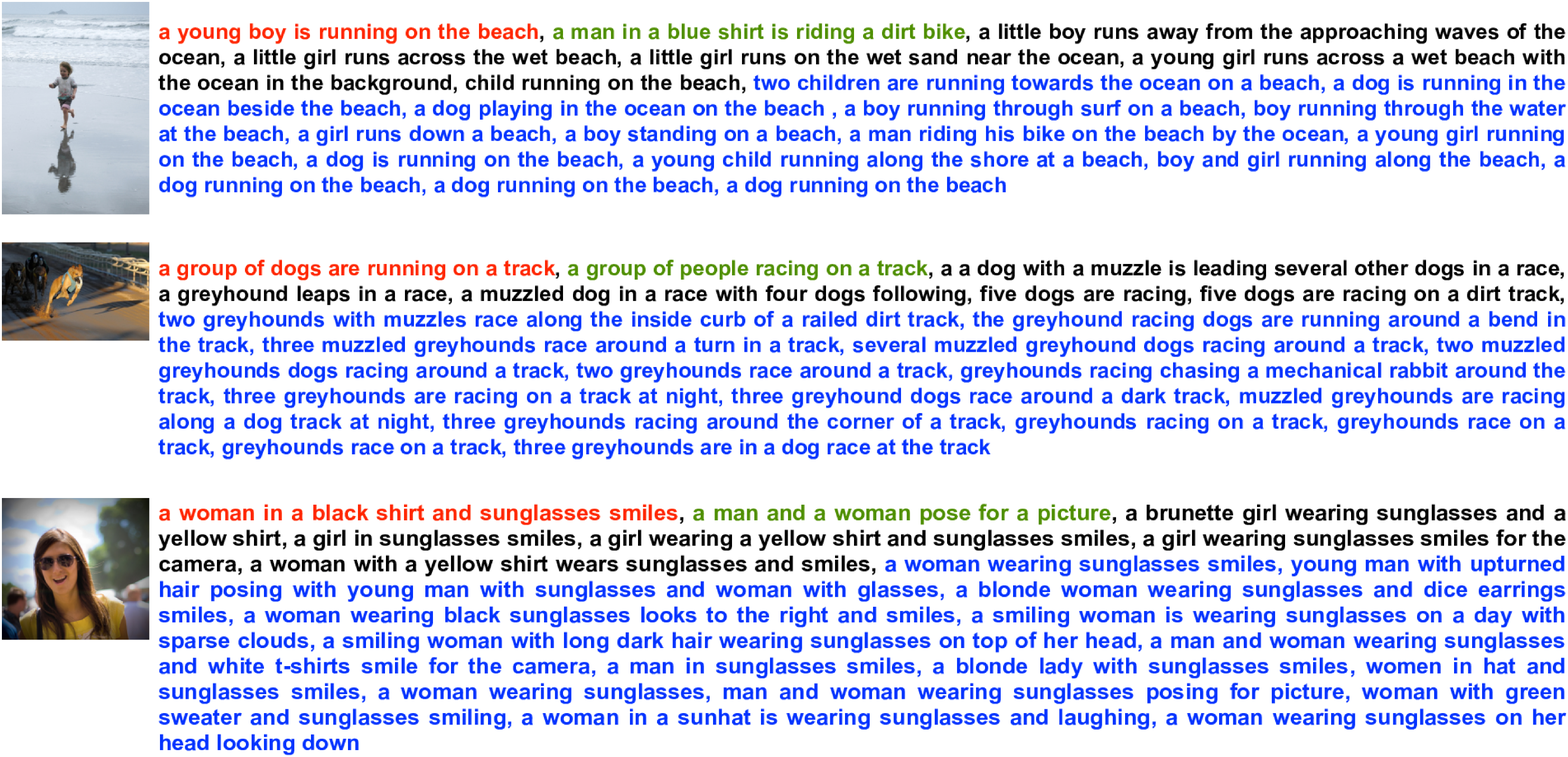}
\vspace{-15mm}
\caption{Results of the LSTM and gLSTM model. 
We mark the generated sentence by LSTM and gLSTM respectively in green and red, 
the ground truth references in black and
the most relevant retrieval results in blue. 
We observe that the retrieval results are helpful to caption generation.
Notice that for the third example, the result of our model is not that accurately
but still much better than the one of the LSTM model.}
\label{fig:examples}
\vspace{-3mm}
\end{figure*}

\noindent\textbf{Evaluation measures.} 
Here we use the two most popular measures in the machine translation and image caption generation literature, 
namely the \emph{BLEU}~\cite{bleu02} and the \emph{METEOR}~\cite{meteor14} measure.

\emph{BLEU} is a precision-based metric.
The main component of BLEU is n-gram precision of the generated caption with respect to the references.
Precision is computed separately for each n-gram and then B@n is computed as a geometric mean of these precisions.
\emph{BLEU} of high order n-gram indirectly measures the grammatical coherence.

However, \emph{BLEU} is criticized to favor short sentences. 
It only considers precision but does not take recall into consideration.
For this reason \emph{METEOR} is also reported in recent works~\cite{ChenZ14a, Fang14, Xu15}. 
\emph{METEOR} evaluates a generated sentence by computing a score based on word level matches 
between the generation and a reference and returning the maximum score over a set of references.
In the computation of the matching score, it considers unigram-precision, unigram-recall and a measure of alignment.
Hence, \emph{METEOR} accounts for precision, recall and the importance of grammaticality.
In user evaluation studies \emph{METEOR}~\cite{meteor07} has been shown to 
have a higher correlation with human judgments than any order of \emph{BLEU}.\\

All scores are computed using the coco-caption code \footnote{\url{https://github.com/tylin/coco-caption}}. 

\noindent\textbf{Implementation details.} 
In the following experiments we use the MatConvNet toolbox~\cite{MatConvNet} and 
the 16-layer OxfordNet~\cite{Simonyan14a} pretrained model 
to compute $CNN$ features and extract the last fully-connected layer's output as image representation.
As for preprocessing of texts, for the neural language model, 
we use the publicly available data where texts are converted to lowercase, non-alphanumeric characters are ignored 
and only words appearing at least 5 times in the training set are kept to create a vocabulary.
For CCA, we use the NLTK toolbox\cite{nltk} to further lemmatize words and 
build a vocabulary based on these words 
(3000 words for flickr8k and 5000 for flickr30k and MS COCO).
Then tf-idf-weighted BoW vectors are computed as sentence representation for CCA.
For Flickr8k and Flickr30K we set the number of dimensions for the image and word embeddings and the hidden layer of the gLSTM to 256. For MSCOCO we set the number to 512
(note that this is much smaller than the one used in other work).
The gLSTM Models are trained with RMSProp ~\cite{rmsprop12}, which is a stochastic gradient descent method 
using an adaptive learning rate algorithm. 
The learning rate is initialized with 1e-4 for Flickr8k and Flickr30k and 4e-4 for MS COCO.
We use dropout and early stopping to avoid overfitting and use validation set log-likelihood for model selection.
For CCA, we set $p=4$ as suggested in ~\cite{GongKIL14} and the dimension of the common space to 200 for Flickr8k and Flickr30k, and 500 for MS COCO
which we find works well in practice.
At the test stage, we set the beam size to 10 for all experiments.
We built our code for the proposed gLSTM model on Karpathy's NeuralTalk code
\footnote{\url{https://github.com/karpathy/neuraltalk}},
which implements the single model in Google's paper~\cite{Vinyals14}.
Note that we take that model as the \textbf{baseline}.

\subsection{Length Normalization}
\label{sec:exp_norm}
\begin{table}[t]
  \centering
  \setlength{\tabcolsep}{3pt}
  \scalebox{1}{
  
  \begin{tabular}{l c c c c c c}
  \toprule
  Normalization & B@1 & B@2 & B@3 & B@4 & METEOR \\
  \midrule
\textit{Baseline} &  59.6 & 40.4 & 26.1 & 17.0 & 17.45 \\
\textit{Polynomial} & 57.8 & 39.2 & 26.0 & 17.6 & \textbf{18.86} \\
\textit{Min-hinge} & 60.4 & 41.4 & 27.6 & \textbf{18.6} & 18.53 \\
\textit{Max-hinge} & 57.6 & 38.8 & 25.2 & 16.7 & 17.65 \\
\textit{Gaussian} & \textbf{60.7} & \textbf{41.7} & \textbf{27.8} & \textbf{18.6} & 18.35 \\
  \bottomrule
  \end{tabular}
  }
  \caption{The performance of different length normalization strategies on Flickr8k.}
\label{tab:perf-sentence-length}
\vspace{-5mm}
\end{table}

In this experiment we evaluate the importance of the sentence length normalization to caption generation. 
We carry out the experiment on the Flickr8k dataset and report the results in Table~\ref{tab:perf-sentence-length}.
%
For clarity we perform this experiment based on the LSTM baseline, not gLSTM.

We observe that compared to the baseline whose selection is based on unnormalized log-likelihood, 
length normalization has a positive effect on either the \emph{BLEU} metric or \emph{METEOR} metric. 
Polynomial, min-hinge and Gaussian normalization respectively bring the largest improvement 
to \emph{METEOR} and \emph{BLEU}.
Therefore, in the following experiments, 
we only report the performance of the proposed gLSTM with these three kinds of length normalization.
Besides, we also compute the average length of generated sentences and references.


 \begin{table}[t]
   \centering
   \setlength{\tabcolsep}{6pt}
   \scalebox{0.75}{
   \small{
   \begin{tabular}{c c c c c c}
   \toprule  
 	\textit{GT Refs} & \textit{Baseline} 	& \textit{Polynom.} 	& \textit{Min-hinge} & \textit{Max-hinge}  & \textit{Gaussian} \\
 	\midrule
 	$10.87 (3.74)$ & $8.75 (2.44)$	& $11.07 (2.62)$ 	& $9.64 (1.92)$ & $9.55 (1.69)$ & $9.57 (3.30)$ \\
   \bottomrule
   \end{tabular}
   }
   }
   \caption{The average and the standard deviation of the sentence length for the ground truth references, 
  and different normalization strategies on Flickr8k. }
  \vspace{-5mm}
 \label{tab:sentence-length}
 \end{table}


\subsection{gLSTM with Different Types of Guidance}

\begin{table}[t]
  \centering
  \setlength{\tabcolsep}{3pt}
  \scalebox{0.90}{
  {\small
  \begin{tabular}{l c c c c c}
  \toprule  
  & B@1 &B@2 &B@3 &B@4 & METEOR \\
  \midrule
  \textit{Baseline, Original}               & 59.6 & 40.4 & 26.1 & 17.0 & 17.45 \\
  \textit{Baseline, Polynomial}               & 57.8 & 39.2 & 26.0 & 17.6 & 18.86 \\  
  \textit{Baseline, Min-hinge}               & 60.4 & 41.4 & 27.6 & 18.6 & 18.53 \\  
  \textit{Baseline, Gaussian}				& 60.7 & 41.7 & 27.8 & 18.6 & 18.35 \\
  \midrule 
  \textit{Baseline 512, Original} & 61.0 & 42.4 & 28.6 & 18.9 & 18.21 \\
  \textit{Baseline 512, Polynomial} & 58.2 & 40.2 & 27.1 & 18.1 & 19.83 \\  
  \textit{Baseline 512, Min-hinge} & 61.3 & 42.9 & 29.2 & 19.6 & 19.13 \\
  \textit{Baseline 512, Gaussian} & 61.3 & 42.8 & 29.1 & 19.5 & 19.07 \\
  \midrule
  \textit{ret-gLSTM, Original} & 63.4 & 43.7 & 29.2 &  19.3 & 18.54 \\
  \textit{ret-gLSTM, Polynomial} & 58.8 & 40.4 & 27.5 & 18.6 & 19.86 \\
  \textit{ret-gLSTM, Min-hinge} & 63.0 & 43.8 & 29.9 & 20.2 & 19.46 \\
  \textit{ret-gLSTM, Gaussian} & 63.5 & 44.2 & 30.2 & 20.6 & 19.38 \\
  \midrule
  \textit{emb-gLSTM, Original} & 63.7 & 44.7 & 30.2 & 20.2 & 19.10 \\
  \textit{emb-gLSTM, Polynomial} & 61.0 & 43.0 & 29.6 & 20.1 & \textbf{20.60} \\
  \textit{emb-gLSTM, Min-hinge}  & 64.3 & 45.7 & 31.6 & 21.5 & 20.28 \\
  \textit{emb-gLSTM, Gaussian} & \textbf{64.7} & \textbf{45.9} & \textbf{31.8} & \textbf{21.6} & 20.19 \\
  \midrule
  \textit{img-gLSTM, Original} & 61.5 & 42.5 & 27.2 & 16.7 & 17.10 \\
  \textit{img-gLSTM, Polynomial} & 55.7 & 38.1 & 24.9 & 15.8 & 17.69 \\
  \textit{img-gLSTM, Min-hinge} &  60.4 & 41.9 & 27.6 & 17.7 & 17.76 \\
  \textit{img-gLSTM, Gaussian} &  60.1 & 41.4 & 27.2 & 17.3 & 17.69 \\
  \bottomrule
  \end{tabular}
  }
  }
  \caption{Comparison between gLSTM with different semantic information on Flickr8k.  
  We denote the gLSTM model with retrieval-based guidance as ret-gLSTM, 
  the one with semantic embedding guidance as emb-gLSTM,
  and the one with image as guidance as img-gLSTM.}
  \vspace{-5mm}
\label{tab:clstm}
\end{table}

\begin{table*}[t]
  \centering
  \setlength{\tabcolsep}{3pt}
  \scalebox{0.90}{
  {\small
  \begin{tabular}{l c c c c c c c c c c c}
  \toprule  
& \multicolumn{5}{c}{\textit{Flickr8k}} & & \multicolumn{5}{c}{\textit{Flickr30k}} \\
  \midrule
  & B@1 &B@2 & B@3 & B@4 & METEOR & & B@1 &B@2 &B@3 &B@4 & METEOR \\
  \cmidrule{2-6} \cmidrule{8-12} 
  \textit{LogBilinear~\cite{KirosICML14}} & 65.6  & 42.4  & 27.7 & 17.7 & 17.31 & & 60.0 & 38.- & 25.4 & 17.1 & 16.88 \\
  \textit{multimodal RNN~\cite{Karpathy14}} & 57.9 & 38.3 & 24.5 & 16.0 & 16.7 & & 57.3 & 36.9 & 24.0 & 15.7 & 15.3 \\
  \textit{Google NIC~\cite{Vinyals14}}         & 63.- & 41.- & 27.- & \textemdash & \textemdash & & 66.3 & 42.3 & 27.7 & 18.3 & \textemdash \\  
  \textit{LRCN-CaffeNet~\cite{Donahue14}}        & \textemdash & \textemdash & \textemdash & \textemdash & \textemdash & &  58.7 & 39.1 & 25.1 & 16.5 & \textemdash \\
  \textit{m-RNN-AlexNet~\cite{Mao14}}        & \textemdash & \textemdash & \textemdash & \textemdash & \textemdash & & 54.- & 36.- & 23.- & 15.- & \textemdash \\
  \textit{m-RNN~\cite{Mao14}}       & \textemdash & \textemdash & \textemdash & \textemdash & \textemdash & & 60.- & 41.- & 28.- & 19.- & \textemdash \\
  \textit{Soft-Attention~\cite{Xu15}}         & \textbf{67.-} & 44.8 & 29.9 & 19.5 & 18.93 & & 66.7 & 43.4 & 28.8 & 19.1 & 18.49 \\
  \textit{Hard-Attention~\cite{Xu15}}         & \textbf{67.-} & 45.7 & 31.4 & 21.3 & 20.3 & & \textbf{66.9} & 43.9 & 29.6 & 19.9 & 18.46 \\
  \midrule
  \textit{emb-gLSTM, Polynomial}   & 61.0 & 43.0 & 29.6 & 20.1 & \textbf{20.60} & & 59.8 & 41.3 & 29.3 & 19.2 & \textbf{18.58} \\
  \textit{emb-gLSTM, Min-hinge}  & 64.3 & 45.7 & 31.6 &21.5 & 20.28 & & 63.8 & 44.1 & 30.2 & 20.5 & 18.13 \\
  \textit{emb-gLSTM, Gaussian}  & 64.7 & \textbf{45.9} & \textbf{31.8} & \textbf{21.6} & 20.19 & & 64.6 & \textbf{44.6} & \textbf{30.5} & \textbf{20.6} & 17.91 \\
  \bottomrule
  \end{tabular}
  }
  }
  \caption{Comparison with state-of-the-art methods on Flickr8k and Flickr30k. }
\label{tab:state-of-the-art-generation}
\vspace{-5mm}
\end{table*}

In this experiment we evaluate the gLSTM model with different types of semantic information, 
as described in Section~\ref{sec:contextual_info}.
For fair comparison, we also apply beam search with length normalization to the baseline.
We run this experiment on Flickr8k and report the results in Table~\ref{tab:clstm}.

The result illustrates that semantic information brings much improvement in the performance,
especially emb-gLSTM, the gLSTM with semantic embedding guidance.
We also observe that img-gLSTM, the gLSTM with the image itself as guidance, 
does not bring any improvement but even deteriorates the performance.
Besides, we also conduct an experiment for a baseline but with more parameters (512 dimension instead of 256 dimension) for each gate to emphasize the improvement mainly comes from the global guide. 
The total number of network parameters is 5.2M in total compared to 5.9M and 3.1M for the proposed gLSTM ret-gLSTM and B. 
As is shown in Table~\ref{tab:clstm}, we can see that increasing parameters indeed improves the performance, 
but still a little worse than the proposed emb-gLSTM even though it has much fewer parameters.

\subsection{Comparison with State-of-the-art methods}

We compare the proposed gLSTM with state-of-the-art methods for caption generation in the literature. 
We perform the experiment on Flickr8k and Flickr30k and report the results in Table~\ref{tab:state-of-the-art-generation}.
We only evaluate emb-gLSTM in this experiment,
since it is computationally efficient and obtains the best performance among different models in the previous experiment.
For most evaluated methods, they use CNN with deeper network architecture 
such as OxfordNet~\cite{Simonyan14a} and GoogLeNet~\cite{Szegedy14}.
Methods which do not use a deeper CNN include LRCN-CaffeNet~\cite{Donahue14} and m-RNN-AlexNet~\cite{Mao14}.
Note that Google's method~\cite{Vinyals14} uses an ensemble of multiple LSTM models, while ours only uses a single emb-gLSTM model.
We can see from the table, the proposed emb-gLSTM model performs favorably against state-of-the-art approaches.
Interestingly, it performs even on par with the latest state-of-the-art~\cite{Xu15}, 
which is based on more complicated and expensive attention mechanisms. 

\begin{table}[htbp]
  \centering
  \setlength{\tabcolsep}{3pt}
  \scalebox{0.90}{
  {\small
  \begin{tabular}{l c c c c c c}
  \toprule    
  & B@1 & B@2 & B@3 & B@4 & METEOR & mCIDEr\\
  \midrule
  \textit{multimodal RNN~\cite{Karpathy14}} & 62.5 & 45.0 & 32.1 & 23.0 & 19.5 & 66\\
  \textit{Google NIC~\cite{Vinyals14}}         & 66.6 & 46.1 & 32.9 & 24.6 &  \textemdash & \textemdash\\  
  \textit{LRCN-CaffeNet~\cite{Donahue14}}        & 62.8 & 44.2 & 30.4 & \textemdash & \textemdash & \textemdash \\
  \textit{m-RNN~\cite{Mao14}}       & 67 & 49 & 35 & 25 & \textemdash & \textemdash\\
  \textit{Soft-Attention~\cite{Xu15}}         & 70.7 & 49.2 & 34.4 & 24.3 & \textbf{23.9} & \textemdash\\
  \textit{Hard-Attention~\cite{Xu15}}         & \textbf{71.8} & \textbf{50.4} & 35.7 & 25.0 & 23.04 & \textemdash\\
  \midrule
  \textit{emb-gLSTM, Polynomial}   & 63.8 & 46.3 & 33.6 & 24.8 & 23.33 & 79.03\\
  \textit{emb-gLSTM, Min-hinge}  & 66.3 & 48.5 & 35.4 &26.2 & 22.95 & \textbf{81.26}\\
  \textit{emb-gLSTM, Gaussian}  & 67.0 & 49.1 & \textbf{35.8} & \textbf{26.4} & 22.74 & 81.25 \\
  \bottomrule
  \end{tabular}
  }
  }
  \caption{Comparison with state-of-the-art methods on MS COCO. }
\label{tab:state-of-the-art-generation-COCO}
\vspace{-5mm}
\end{table}

\vspace{-1mm}
\section{Conclusion}
In this work we have proposed an extension of the LSTM model for image caption generation.
By adding semantic information as extra input to each unit of the LSTM block,
we have shown that the model can better stay ``on track", describing the image content
without drifting away to unrelated yet common phrases.
In addition, we explore different types of length normalization for beam search
in order to prevent a bias towards very short sentences, which further improves the results.
The proposed method achieves state-of-the-art performance on various benchmark datasets.
Moreover, our key contributions are, to a large extent, complementary to key aspects of other methods,
uch as attention mechanisms~\cite{Xu15} or model ensembles~\cite{Vinyals14},
indicating that further improvements on performance may be obtained by integrating these schemes.

\vspace{-1mm}
\section{Acknowledgment}
The authors acknowledge the support of the IWT-SBO project PARIS.

{\small
\bibliographystyle{ieee}
\bibliography{clstm}
}

\end{document}